\documentclass{article}
\usepackage{ijcai20}

\usepackage{makecell}
\usepackage{times}
\usepackage{soul}
\usepackage{url}
\usepackage[hidelinks]{hyperref}
\usepackage[utf8]{inputenc}
\usepackage[small]{caption}
\usepackage{graphicx}
\usepackage{amsmath}
\usepackage{amsthm}
\usepackage{booktabs}
\usepackage{algorithm}
\usepackage{algorithmic}
\title{2nd Place Solution to Instance Segmentation of IJCAI 3D AI Challenge 2020}

\author{
	Kai Jiang$^1$\footnote{Equal Contribution}\and
	Xiangyue Liu$^2$\footnotemark[1]\and
	Zheng Ju$^3$\footnotemark[1]\And
	Xiang Luo$^1$\\
	\affiliations
	$^1$LinkDoc Technology, Beijing, China\\
	$^2$School of Software, Beihang University, Beijing, China\\
	$^3$Huaxin consulting Co.,Ltd, Hangzhou, China\\
	\emails
	\{tftyjk, starcxlxy\}@hotmail.com,
	uki\_tuantuan@hotmail.com,
	xiangluo1019@gmail.com
}

\begin{document}
	
	\maketitle
	
	\begin{abstract}	
		Compared with MS-COCO, the dataset for the competition has a larger proportion of large objects which area is greater than 96$\times$96 pixels. As getting fine boundaries is vitally important for large object segmentation, Mask R-CNN with PointRend is selected as the base segmentation framework to output high-quality object boundaries. Besides, a better engine that integrates ResNeSt, FPN and DCNv2, and a range of effective tricks that including multi-scale training and test time augmentation are applied to improve segmentation performance. Our best performance is an ensemble of four models (three PointRend-based models and SOLOv2), which won the 2nd place in IJCAI-PRICAI 3D AI Challenge 2020: Instance Segmentation.
	\end{abstract}
	
	\section{Introduction}
	Follow as object detection, instance segmentation can be grouped into two genres: two-stage methods (e.g., Mask R-CNN~\cite{2017arXiv170306870H} and its variants) and one-stage methods (e.g., YOLACT~\cite{Bolya_2019_ICCV} and its variants). The former generates proposals in the first stage and the classification, fine location regression and mask are predicted in the second stage, while the later predicts bounding boxes, class and mask simultaneously. In general, one-stage methods are faster but two-stage methods are more accurate.
	
	The images for segmentation are from a large-scale indoor dataset named 3D FUTURE~\cite{fu20203dfuture}, and it has two characteristics. First, it has a large percentage of large objects, which indicates the detection is not the bottleneck and the classification of pixels around object boundaries is important. Second, the large class imbalance of the datasets may prevent model fitting.
	
	To tackle these difficulties, a two-stage method, Mask R-CNN with PointRend~\cite{2019arXiv191208193K} (denoted PointRend), is adopted to refine object boundaries,  and the focal loss~\cite{2017arXiv170802002L} is applied to alleviate the class imbalance. An optimized backbone with combining ResNeSt~\cite{2020arXiv200408955Z}, FPN~\cite{2016arXiv161203144L} and DCNv2~\cite{2018arXiv181111168Z} enables the networks to capture complex features. A conservative data augmentation is used to combat over-fitting. With the help of multi-scale training, test time augmentation and model ensemble, we achieve the best mAP of 78.7 on trackA and 77.2 on trackB.
	
\begin{figure}[!h]
	\centering
	\includegraphics[scale=0.55]{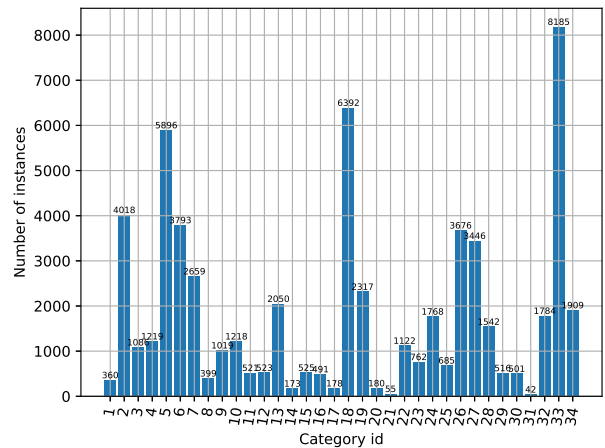}
	\caption{\textbf{The number of instances for 34 categories.}}
	\label{fige1}
\end{figure}

\section{Datesets}
\subsection{Data Analysis}

The dataset in the competition includes 12,144, 2,024, 6,072 images in its training, validation, and test set, respectively, and there are 34 categories. The size of each picture is 1200$\times$1200 pixels. We analyze the dataset from three aspects: (1) category distribution, (2) the area of objects, and (3) aspect ratio of objects.

\textbf{Category distribution.} Figure 1 indicates that the network faces a large class imbalance, which will lead to degenerate models during training. A common solution is to perform some forms of resampling or reweighing. In our experiments, we alleviate the class imbalance by focal loss.

\textbf{Area of objects.} The objects with an area greater than 96$\times$96 pixels and 256$\times$256 pixels account for over 80\% and 25\% of the total 61,010 instances in the training set, respectively. In general, the larger the object, the more accurate the detector, so the detection task in the competition is not a bottleneck. In other words, a good segmentation becomes vitally important for winning the competition. Most methods can classify pixels inside the object accurately but predict pixels around boundaries unsatisfactorily as objects have irregular boundaries and pixels around edges are hard samples. PointRend samples these hard pixels and classify them by MLP to output high-quality object boundaries, which attracts us to adopt it as the base framework.

\textbf{Aspect ratio.} The anchor size is set from 64 to 1024 pixels with three aspect ratios of {1:2, 1:1, 2:1} and there are two reasons: (1) The image is scaled to a size selected randomly from 1200 to 1500 pixels before feeding into network, which instructs us in anchor size setting. (2) The aspect ratio of all instances in the training set is counted and aspect ratios of {1:2, 1:1, 2:1} is proper.

\subsection{Data Preprocessing}

\textbf{Conservative Image Augmentation.} We adopt a conservative image augmentation to alleviate the over-fitting. The images are randomly left-right flipped and changed brightness, contrast, and saturability with a small ratio. The experiments demonstrate the aggressive augmentation, such as Gaussian blur or random rotation, causes mAP reduction.

\textbf{Mask Correct.} As shown in Figure 2, some unexplainable pixel labels in mask are observed during the data visualization. We think these mislabeled pixels will cause two problems: (1) Training will be unstable due to wrong labels, (2) The mask fitting will be affected by the noises. Therefore, masks are corrected before inputting the net, which can achieve an improvement of 0.98 mask mAP from 73.40 to 74.38 (Table 2).

\begin{figure}[htbp]
	\centering
	\includegraphics[width=0.9\linewidth]{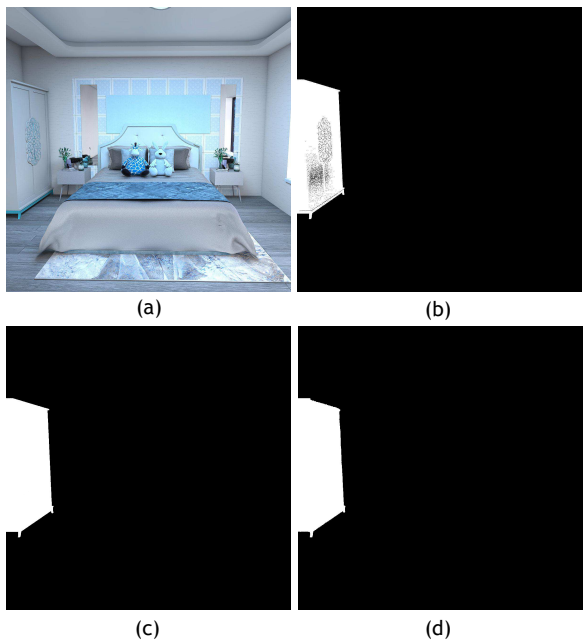}
	\caption{\textbf{Example of mask correct.} (a) Original image. (b) Original mask. (c) Corrected mask. (d) mask prediction.}
\end{figure}

\section{Method}

\subsection{Network}

An enhanced PointRend is trained to segment furniture from images. Its backbone is a fusion of ResNeSt, FPN and DCNv2, and this strong engine is responsible for complex features. A cascade technique is used to capture better proposal features. This segmentor is called \textbf{SPR-Net} for short, where S stands for "Super". If not specified, all mAP mentioned below are achieved on the validation set.

The SPR-Net consists of a backbone, RPN layer and final prediction layer, and its structure is shown in Figure 3. The backbone is a ResNet~\cite{2015arXiv151203385H} variant, named \textbf{ResNeSt}, and it integrates ideas of group convolution proposed by GoogleNet~\cite{2014arXiv1409.4842S} and group-attention proposed by SK-Net~\cite{Li_2019_CVPR}. ResNeSt outperforms all existing ResNet variants and achieves state-of-the-art performance on almost all CV tasks, including image classification, object detection and instance segmentation. In our experiments, PointRend with ResNeSt101 achieves a 4.11 mAP and 1.79 mAP improvement compared to ResNet101 and ResNeXt101(Table 1), respectively. Besides, we find that models utilized ResNeSt200 obtain a worse performance, which may be caused by vanishing of localization information due to deeper layers and overfitting because of more parameters and limited images.

\textbf{Feature Pyramid Network (FPN)} is utilized to capture multi-level information, which is essential because the size and category of objects have large variations. In general, features in deep layers will have high-level semantic information that is beneficial to category recognition, and features in shallower layers help localization as it contains more information about edges and contours. In this work, if not specified, FPN is a default setting for all models. Another effective technique is \textbf{Deformable Convolution Network (DCN)}~\cite{2017arXiv170306211D}. By adding offsets to convolution, DCN learns geometric transformation of objects for finer features. This method applied to the backbone of SPR-Net is an improvement version (\textbf{DCNv2}), and it not only offsets the input but also adjusts the weight of the input at each position. SPR-Net with DCNv2 achieves 72.76 mAP on the validation set, which surpasses that without DCNv2 by 0.94. In the RPN layer of SPR-Net, the \textbf{cascade technique}, which learns proposals from a coarse to fine locations, is used to learn more accurate proposals. In addition, high-quality proposals are beneficial to hard sample mining since wrong proposals will be regarded as hard samples to pay more attention. On the dataset (1200 images) split randomly from the training set, cascaded PointRend with ResNeSt101 can give an improvement of about 1.0 mAP. Same as FPN, cascade technique is also a default setting for the PointRend-based models.

\begin{table}[h]
	\centering
	\begin{tabular}{l|l|l}
		\hline
		\makecell[c]{Model} & {Backbone}  & {mAP} \\
		\hline
		\makecell[c]{Mask R-CNN} & {ResNet50}  & {52.10} \\
		\hline
		& {ResNet50}   & \makecell[c]{66.75} \\
		\makecell[c]{PointRend} & {ResNet101}  & \makecell[c]{67.71} \\
		& {ResNeXt101}  & \makecell[c]{70.03} \\
		& {ResNeSt101}  & \makecell[c]{\textbf{71.82}} \\
		\hline
	\end{tabular}
	\caption{\textbf{Performance of Mask R-CNN with different backbone and mask head on the validation set.} Note that FPN is a default set.}
\end{table}

As analyzed above, a precise segmentation is vitally important when detection is not a bottleneck. The instance segmentation methods based on the Mask R-CNN usually have a fine performance, but these region-based architectures typically predict masks on a fixed size grid such as 28$\times$28. For large objects in 3D FUTURE, this small size grid produces undesirable ``blobby'' output mask that ignores details of object boundaries (Figure 4, images in top row). \textbf{PointRend} alleviates this problem by a direct and effective approach that reclassifies these hard edge pixels. As shown in Figure 3, it yields a coarse mask prediction (size is 7$\times$7 or larger) for each object firstly, then it samples a set of uncertain points (probabilities closest to 0.5) from coarse mask and predicts the label of these points by an MLP classifier. The feature input to MLP is from two sources: (1) a fine-grained feature map of the proposals, (2) the coarse prediction mask. Visually, compare to Mask R-CNN, PointRend can predict masks with substantially finer details around object boundaries (Figure 4, images in bottom row). In terms of quantitative improvement, using ResNet50 with FPN, Mask R-CNN with PointRend achieves a great improvement of 14.65 mAP compared to that with standard mask head from 52.10 to 66.75 (Table 1).

\begin{figure}[htbp]
	\centering
	\includegraphics[width=1.0\linewidth]{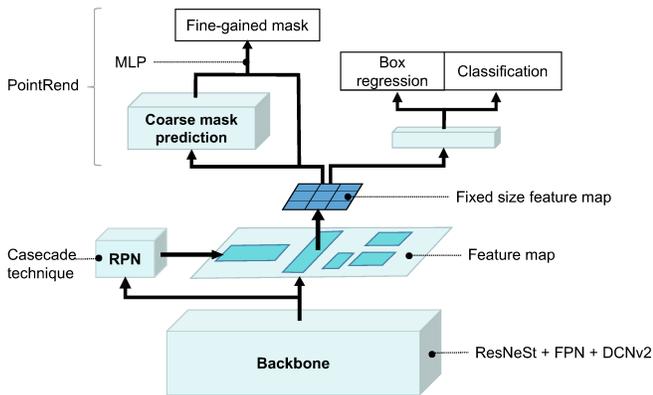}
	\caption{\textbf{The structure of SPR-Net.}}
\end{figure}

\subsection{Loss Function}

The loss of SPR-Net consists of classification loss, bounding box regression loss, mask loss and mask point loss. In our scheme, classification loss and mask loss are set to focal loss to alleviate the class imbalance and mine the hard samples.

\textbf{Focal Loss} is a commonly used technique in object detection. It is defined as:
$$ FL(p_{t}) = -\alpha _{t}(1-p_{t})^{r}log(p_{t})\eqno(1)$$where ${\alpha _{t}}$ denotes the weight of classification loss that sample is predicted to class ${t}$, and ${r}$ smoothly adjusts the rate at which easy examples are down-weighted.

As mentioned above, the dataset has a large class imbalance. In SPR-Net, the cross-entropy loss for classification is replaced by a \textbf{multi-class focal loss}.  The categories with a poor mAP and a small number are up-weighted by set a larger ${\alpha _{t}}$. Similarly, a \textbf{mask focal loss} is applied to mask branch and all ${\alpha}$ is set to 1, ${r}$ of multi-class focal loss and mask focal loss is set to 2. To drive model learning segmentation task better, the weight of mask loss is set to 1.1 and that of the others is set to 1. Finally, SPR-Net equipped with focal loss gives a gain of 0.66 mAP (Table 2).

\begin{figure}[htp]
	\centering
	\includegraphics[width=1.0\linewidth]{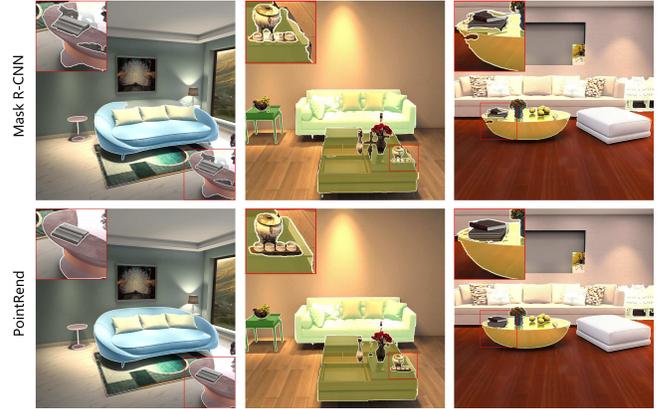}
	\caption{\textbf{Examples of segmentation result from Mask R-CNN with default mask head vs. with PointRend, using ResNet101 with FPN.}}
\end{figure}

\begin{table*}[h]
	\centering
	\begin{tabular}{l|lllllll|l|lllll}
		\hline
		\makecell[c] & {X101}& {S101}  & {DCNv2} & {Focal} & {MC} & {MST}  &{TTA} & {mAP} &AP50 &AP75 &APs &APm &APl\\
		\hline
		&\makecell[c]{$\surd$} &    &{}    & &  & & &70.32&87.49&75.30&43.71&64.17&78.30 \\
		&& \makecell[c]{$\surd$} & & &  &  & &71.82&88.09&77.04&49.01&67.27&79.28 \\
		\makecell[c]{PointRend+} && \makecell[c]{$\surd$}  &\makecell[c]{$\surd$} & &  & & &72.76&88.39&78.07&52.03&66.61&79.55\\
		\makecell[c]{Cascade+} && \makecell[c]{$\surd$}  &\makecell[c]{$\surd$} &\makecell[c]{$\surd$} &&& &73.40&88.71&78.52&52.63&66.76&79.65\\
		\makecell[c]{FPN} &&\makecell[c]{$\surd$} &\makecell[c]{$\surd$} &\makecell[c]{$\surd$}&\makecell[c]{$\surd$}&& &74.38&89.00&79.62&54.46&66.54&79.69\\
		&& \makecell[c]{$\surd$}  &\makecell[c]{$\surd$} &\makecell[c]{$\surd$} &\makecell[c]{$\surd$}&\makecell[c]{$\surd$} &&75.53&89.92&80.55&54.68&67.36&81.63\\
		&& \makecell[c]{$\surd$}  &\makecell[c]{$\surd$} &\makecell[c]{$\surd$}
		&\makecell[c]{$\surd$}&\makecell[c]{$\surd$}&\makecell[c]{$\surd$} &\textbf{77.12}&\textbf{90.46}&\textbf{82.70}&\textbf{57.42}&\textbf{71.60}&\textbf{81.80}\\
		\hline
	\end{tabular}
	\caption{\textbf{PointRend's gradual performance improvement on the validation set.} ``X101'' denotes ResNetXt101, ``S101'' denotes ResNeSt101, ``DCNv2'' denotes Deformable Convolution Network v2, ``Focal'' denotes Focal Loss, ``MC'' denotes Mask Correct, ``MST'' denotes Multi-Scale Training, ``TTA'' denotes Test Time Augmentation.}
\end{table*}


\section{Tricks}

Several tricks, including multi-scale training, test time augmentation and model ensemble, are utilized for better performance.

\textbf{Multi-Scale Training.} The best way to detect and segment objects is to enlarge them. Compared to single-scale training (1200 pixels), the multi-scale training (1200-1500 pixels) with the same net achieves a 1.15-point mask mAP gap
(74.38 vs. 75.53).

\textbf{Test Time Augmentation.} In inference phase, the test images with different transformations (HorizontalFlip and Multi-Scale Test) are feed to the trained model, and then compute the average results of them to get the more confident answer. This method can give an improvement of 1.59 mask mAP from 75.53 to 77.12.

\textbf{Model ensemble} is a classic and effective technique for robust performance. In our scheme, the model bags for ensemble have three PointRend-based models and a \textbf{SOLOv2}\cite{2020arXiv200310152W}, which is trained to increase diversity between models. SOLOv2 is a bottom-up method, which segments objects directly rather than depending on box detection. Using ResNetXt101 with DCNv2 and FPN, SOLOv2 achieves a mask mAP 75.50 on the validation set (Table 3).

Finally, model ensemble achieves the best performance in the competition, which mAP is 78.7 on trackA and 77.2 on trackB, by averaging predictions of models in model bags.

\begin{table*}[h]
	\centering
	\begin{tabular}{lllll|l|l}  
		\hline
		Method & Backbone &  & Loss & Scale & mAP (Valid) & mAP (Test) \\ 
		\hline
		Mask R-CNN & X101 &  & Cross Entropy & (1200, 1500) & 74.30& 73.10 \\
		Cascade R-CNN & ResNeSt101 &  & Cross Entropy & (1200, 1500) & 76.64 & 74.44 \\
		Cascade R-CNN & ResNeSt101 &  & Focal Loss & (1200, 1500) & 77.12 & 75.86 \\
		SOLOv2 & ResNetXt101 &  & Cross Entropy  & (1200, 1600) & 75.50 & 74.61 \\
		\hline
		Ensemable &  &  &  & & \textbf{78.75 (traceA)} & \textbf{77.22 (traceA)} \\
		\hline
	\end{tabular}
	\caption{\textbf{The details of model ensemble.} ``Valid'' denotes the Validation set, ``Test'' denotes the Test set. The models based on Mask R-CNN have a mask head with PointRend and all the models have a backbone with FPN and DCNv2.}
	\label{tab:plain}
\end{table*}

\section{Conclusions}
In this report, we present the main details of our scheme that utilizes reasonable data processing, effective models, model ensemble, and other strategies, which gradually increase the leaderboard score step by step, allowing us to achieve the 2nd place in the Instance Segmentation of IJCAI 3D AI Challenge 2020. 
\bibliographystyle{named}
\bibliography{ijcai20}

\end{document}